\documentclass[11pt,a4paper]{article}

\usepackage{arabtex}
\usepackage[hyperref]{acl2017}

\usepackage{acl2017}

\usepackage{times}

\usepackage{amssymb}
\usepackage{amsmath}
\usepackage{array}
\usepackage{color}
\usepackage{dcolumn}
\usepackage{graphicx}
\usepackage{hyperref}
\usepackage{latexsym}
\usepackage{multicol}
\usepackage{multirow}
\usepackage{rotating}
\usepackage{url}

\usepackage{latexsym}

\usepackage{url}

\aclfinalcopy % Uncomment this line for the all SemEval submissions
\title{SemEval-2017 Task 4: Sentiment Analysis in Twitter}

\author{
Sara Rosenthal$^{\clubsuit}$, Noura Farra$^{\diamondsuit}$, Preslav Nakov$^{\heartsuit}$ \\
$^{\heartsuit}$Qatar Computing Research Institute,
Hamad bin Khalifa University, Qatar \\
$^{\diamondsuit}$Department of Computer Science, Columbia University \\
$^{\clubsuit}$IBM Research, USA \\
}

\date{}

\begin{document}
\maketitle
\begin{abstract}
This paper describes the fifth year of the \emph{Sentiment Analysis in Twitter task}. SemEval-2017 Task 4 continues with a rerun of the subtasks of SemEval-2016 Task 4, which include identifying the \textit{overall sentiment} of the tweet, \textit{sentiment towards a topic} with classification on a two-point and on a five-point ordinal scale, and \textit{quantification} of the distribution of sentiment towards a topic across a number of tweets: again on a two-point and on a five-point ordinal scale. Compared to 2016, we made two changes: (\emph{i})~we introduced a new language, Arabic, for all subtasks, and (\emph{ii})~we made available information from the profiles of the Twitter users who posted the target tweets. The task continues to be very popular, with a total of 48 teams participating this year.
\end{abstract}

\section{Introduction}
\label{sec:intro}
The identification of sentiment in text is an important field of study, with social media platforms such as Twitter garnering the interest of researchers in language processing as well as in political and social sciences. The task usually involves detecting whether a piece of text expresses a \textsc{Positive}, a \textsc{Negative}, or a \textsc{Neutral} sentiment; %sometimes, an \textsc{Objective} class is also used.
%NF: Could be confusing as we never mention objective in the paper and we actually conflate it with neutral. For the topic case, we did not have neutral.
the sentiment can be general or about a specific topic, e.g., a person, a product, or an event.

The \emph{Sentiment Analysis in Twitter} task has been run yearly at SemEval since 2013 \cite{Nakov:2013ty,rosenthal-EtAl:2014:SemEval,Nakov:2016rm}, with the 2015 task introducing sentiment towards a topic \cite{rosenthal-EtAl:2015:SemEval} and the 2016 task introducing tweet quantification and five-point ordinal classification \cite{nakov-EtAl:2016:SemEval1}.
% with all four tasks attracting a high number of participating teams.

\noindent SemEval is the International Workshop on Semantic Evaluation, formerly SensEval.
It is an ongoing series of evaluations of computational semantic analysis systems, organized under the umbrella of SIGLEX, the Special Interest Group on the Lexicon of the Association for Computational Linguistics.
Other related tasks at SemEval have explored sentiment analysis of product review and their aspects \cite{pontiki2014semeval,pontiki2015semeval,pontiki2016semeval}, sentiment analysis of figurative language on Twitter \cite{ghosh2015semeval}, implicit event polarity \cite{russo2015semeval}, detecting stance in tweets \cite{mohammad2016semeval}, out-of-context sentiment intensity of words and phrases \cite{kiritchenko2016semeval}, and emotion detection \cite{strapparava2007semeval}. Some of these tasks featured languages other than English, such as Arabic \cite{pontiki2016semeval,mohammad2016semeval}; however, they did not target tweets, nor did they focus on sentiment towards a topic. 
  
This year, we performed a re-run of the subtasks in SemEval-2016 Task 4, which, in addition to the overall sentiment of a tweet, featured classification, ordinal regression, and quantification with respect to a topic. Furthermore, we introduced a new language, Arabic.
%,  for which we collected new training and testing datasets. %Naturally, we also prepared a new testing set for English. 
Finally, we made available to the participants demographic information about the users who posted the tweets, which we extracted from the respective public profiles.
%in Twitter.

\paragraph{Ordinal Classification}
As last year, SemEval-2017 Task 4 includes sentiment analysis on a
  five-point scale \{\textsc{HighlyPositive}, \textsc{Positive},
  \textsc{Neutral}, \textsc{Negative}, \textsc{HighlyNegative}\}, which is in line with product ratings occurring in the corporate world, e.g., Amazon, TripAdvisor, and Yelp.
In machine learning terms, moving from a categorical two-point scale to an ordered five-point scale means moving from binary to \emph{ordinal classification} (aka \textit{ordinal regression}).
  
  %Moving from a categorical two/three-point scale to an ordered
  %five-point scale means, in machine learning terms, moving from
 % binary to \textit{ordinal classification} (a.k.a.\
 % \textit{ordinal regression}).

\paragraph{Tweet Quantification} 
SemEval-2017 Task 4 includes \textit{tweet quantification} tasks along with tweet classification tasks, also on 2-point and 5-point scales. While the tweet classification task is concerned with whether a specific tweet expresses a given sentiment towards a topic, the tweet quantification task looks at estimating the \textit{distribution} of tweets about a given topic across the different sentiment classes. Most (if not all) tweet sentiment classification studies within political science
  \cite{Borge-Holthoefer:2015dz,Kaya:2013ca,marchettibowick-chambers:2012:EACL2012},
  economics \cite{Bollen:2011bf,OConnor:2010fk}, social science
  \cite{Dodds:2011uq}, and market research
  \cite{Burton:2011sh,Qureshi:2013fb}, study Twitter with an interest in aggregate statistics about sentiment and are \emph{not} interested in the sentiment expressed in individual tweets.
%    Estimating the distribution of the classes in a set of unlabelled items by leveraging training data is called \textit{quantification} in data mining and related fields. 
We should also note that quantification is not a mere byproduct of classification, as it can be addressed using different approaches and it also needs different evaluation measures  \cite{Forman:2008kx,Esuli:2015gh}.
  
\paragraph{Analysis in Arabic} 
This year, we added a new language, Arabic, in order to encourage participants to experiment with multilingual and cross-lingual approaches for sentiment analysis. Our objective was to expand the Twitter sentiment analysis resources available to the research community, not only for general multilingual sentiment analysis, but also for multilingual sentiment analysis \emph{towards a topic}, which is still a largely unexplored research direction for many languages and in particular for morphologically complex languages such as Arabic.
%we further added Arabic in order to encourage participants to experiment with cross-lingual approaches, which would leverage the English data.

Arabic has become an emergent language for sentiment analysis, especially as more resources and tools for it have recently become available. 
It is also both interesting and challenging due to its rich morphology and abundance of dialectal use in Twitter. 
Early Arabic studies focused on sentiment analysis in newswire \cite{abdul2011subjectivity,elarnaoty2012machine}, but recently there has been a lot more work on social media, especially Twitter \cite{mourad2013subjectivity,abdul2014samar,refaee2014subjectivity,salameh2015sentiment}, where the challenges of sentiment analysis are compounded by the presence of multiple dialects and orthographical variants, which are frequently used in conjunction with the formal written language. %(Mourad and Darwish. 2013; AbdulMageed and Diab., 2014; Refaee and Reiser., 2014; Mohammad et al., 2015). 
%There have been increasing efforts towards building Arabic sentiment lexicons (AbdulMageed and Diab., 2011; Badaro and al., 2015; Eskander and Rambow., 2015), and Arabic dialectal lexicons aimed towards sentiment analysis in Twitter (Refaee and Reiser., 2014;  Mohammad et al., 2015). 

\noindent Some work studied the utility of machine translation for sentiment analysis of Arabic texts  \cite{salameh2015sentiment,mohammad2016translation,refaee2015benchmarking},
%(Mohammad et al., 2015; Salameh et al., 2015; Refaee and Reiser., 2015),
identification of sentiment holders \cite{elarnaoty2012machine}, and sentiment targets \cite{al2015human,farra2015annotating,farra2017smarties}. We believe that the development of a standard Arabic Twitter dataset for sentiment, and particularly with respect to topics, will encourage further research in this regard.

\paragraph{User Information}
Demographic information in Twitter has been studied and analyzed using network analysis and natural language processing (NLP) techniques ~\cite{mislove2011understanding,nguyen2013old,rosenthal2016social}. Recent work has shown that user information and information from the network can help sentiment analysis in other corpora~\cite{hovy2015demographic} and in Twitter~\cite{volkova2013exploring,DBLP:journals/corr/YangE15}. %Similarly, location has been shown to improve age and demographic detection in Twitter~\cite{DBLP:journals/corr/PavalanathanE15}. 
Thus, this year we encouraged participants to use information from the public profiles of Twitter users such as demographics (e.g.,~age, location) as well as information from the rest of the social network (e.g.,~sentiment of the tweets of friends), with the goal of analyzing the impact of this information on improving sentiment analysis.

The rest of this paper is organized as follows. Section \ref{sec:task} presents in more detail the five subtasks of SemEval-2017 Task 4. 
Section~\ref{sec:datasets} describes the English and the Arabic datasets and how we created them. Section~\ref{sec:eval} introduces and motivates the evaluation measures for each subtask. Section~\ref{sec:results} presents the results of the evaluation and discusses the techniques and the tools that the participants used.
Finally, Section~\ref{sec:conclusion} concludes and points to some possible directions for future work.

\section{Task Definition}
\label{sec:task}

\noindent SemEval-2017 Task 4 consists of five subtasks, each offered for both Arabic and English:
%for each of English and Arabic, with a total of 10 competitions.

\begin{enumerate}
\item \textbf{Subtask A:} % Given a tweet, predict whether it is of
  %positive, negative, or neutral sentiment.
  %Message Polarity Classification: 
  Given a tweet, decide whether it expresses \textsc{Positive}, \textsc{Negative} or \textsc{Neutral} sentiment.

\item \textbf{Subtask B:} Given a tweet and a topic, classify the sentiment conveyed towards that topic on a two-point scale: \textsc{Positive} vs. \textsc{Negative}.

\item \textbf{Subtask C:} Given a tweet and a topic, classify the sentiment conveyed in the tweet towards that topic on a five-point scale: \textsc{StronglyPositive}, \textsc{WeaklyPositive}, \textsc{Neutral}, \textsc{WeaklyNegative}, and \textsc{StronglyNegative}.

\item \textbf{Subtask D:} Given a set of tweets about a topic, estimate the \textit{distribution} of tweets across the \textsc{Positive} and \textsc{Negative} classes.

\item \textbf{Subtask E:} Given a set of tweets about a topic, estimate the \textit{distribution} of tweets across the five classes: \textsc{StronglyPositive}, \textsc{WeaklyPositive}, \textsc{Neutral}, \textsc{WeaklyNegative}, and \textsc{StronglyNegative}.

\end{enumerate}

\begin{table}[tbh]
  \begin{center}
  %  \resizebox{\columnwidth}{!} {
    \begin{tabular}{|c|c|c|c|}
      \hline
     \multicolumn{4}{|c|} {\bf Languages: English and Arabic}  \\ \hline
       & \bf Goal & \bf Granularity & \bf Topic \\ \hline
       A & Classification  & 3-point &  No \\ \hline
        B & Classification & 2-point & Yes \\
        C & Classification & 5-point &  Yes \\ \hline
       D & Quantification & 2-point & Yes \\
       E & Quantification & 5-point &  Yes  \\
   %   & & \multicolumn{2}{c|}{Granularity} \\
      %\cline{3-4}
      %& & Two-point & Five-point \\
      %& & (binary) & (ordinal) \\
      \hline
      %\multirow{2}{*}{\begin{sideways}Goal\end{sideways}} & Classification & Subtask B & Subtask C \\
      %\cline{2-4}

    \end{tabular}
   % }
    \caption{\label{tab:tasks} Summary of the subtasks.}
  \end{center}
\end{table}%

Each subtask is run for both English and Arabic.  Subtask A  has been run in all previous editions of the task and continues to be the most popular one (see section~\ref{sec:results}.) Subtasks B-E have all been run at SemEval-2016 Task 4 \cite{nakov-EtAl:2016:SemEval1}, with variants running in 2015 \cite{rosenthal-EtAl:2015:SemEval}. Table \ref{tab:tasks} shows a summary of the subtasks.

\section{Datasets}
\label{sec:datasets}

Our datasets consist of tweets annotated for sentiment on a 2-point, 3-point, and 5-point scales. We made available to participants all the data from previous years~\cite{nakov-EtAl:2016:SemEval1} for the English training sets, and we collected new training data for Arabic, as well as new test sets for both English and Arabic. The annotation scheme remained the same as last year~\cite{nakov-EtAl:2016:SemEval1}, with the key new contribution being to apply the task and instructions to Arabic as well as providing a script to download basic user information. All annotations were performed on CrowdFlower.
Note that we release all our datasets to the research community to be used freely beyond SemEval. 

%\subsection{Tweet Collection}

\subsection{Tweet Collection}

%The process of tweet collection began by finding suitable topics to annotate. 
We chose English and Arabic topics based on popular current events that were trending on Twitter, both internationally and in specific Arabic-speaking countries, using local and global Twitter trends.\footnote{\url{https://trends24.in/}}  The topics included a range of named entities (e.g., \emph{Donald Trump}, \emph{iPhone}), geopolitical entities (e.g., \emph{Aleppo}, \emph{Palestine}), and other entities (e.g., \emph{Syrian refugees}, \emph{Dakota Access Pipeline}, \emph{Western media}, \emph{gun control}, and \emph{vegetarianism}).  We then used the Twitter API to download tweets, along with corresponding user information, containing mentions of these topics in the specified language. We intentionally chose to use some overlapping topics between the two languages in order to encourage cross-language approaches. 
%for test topics that were popular in both English-speaking and Arabic-speaking regions.  

%TODO Sara: Anything to add about user info?
%Sara: I have added this in a data distribution subsection at the end of this section

We automatically filtered the tweets for duplicates and we removed those for which the bag-of-words cosine similarity exceeded 0.6. We then retained only the topics for which at least 100 tweets remained. The training tweets for Arabic were collected over the period of September-November 2016 and all test tweets were collected over the period of December 2016-January 2017.

For both English and Arabic, 
%the topics in the training and in the test sets did not overlap, i.e., 
the topics for the test dataset were different from those in the training and in the development datasets.
 
\setarab
\novocalize

\begin{table*}[t!]
	\centering
	\resizebox{\textwidth}{!} {
		\begin{tabular}{|p{9cm}|p{1.5cm}|p{5.5cm}|}
			\hline
			\multicolumn{1}{|c|}{\bf Tweet} & \multicolumn{1}{c|}{\bf Overall Sentiment} & \multicolumn{1}{c|}{\bf Topic-level Sentiment} \\
			\hline
%            \hline
			Who are you tomorrow? Will you make me smile or just bring me sorrow? \#HottieOfTheWeek Demi Lovato	& \textsc{Neutral}	& Demi Lovato: \textsc{Positive} \\ \hline
			Saturday without Leeds United is like Sunday dinner it doesn't feel normal at all (Ryan)	& \textsc{WeaklyNegative}	& Leeds United: \textsc{HighlyPositive} \\
			\hline
           Apple releases a new update of its OS & \textsc{Neutral} &	Apple: \textsc{Neutral} \\
			\hline
			%Refugees are facing difficulties & \textsc{Negative}	& Refugees: \textsc{Neutral} \\
%			\hline
		\end{tabular}
	}
	\caption{Some English example annotations that we provided to the annotators.}
	\label{T:annotation-examples}
\end{table*}

\begin{table*}[t!]
	\centering
    \resizebox{\textwidth}{!} {
		\begin{tabular}{|p{9cm}|p{1.5cm}|p{5.5cm}|}
			\hline
			\multicolumn{1}{|c|}{\bf Tweet} & \multicolumn{1}{c|}{\bf Overall Sentiment} & \multicolumn{1}{c|}{\bf Topic-level Sentiment} \\
			\hline
%            \hline
            <'bl t.tlq AlnsxT AltjrybyT AlrAb`T ln.zAm Alt^s.gyl> 
           
			\textit{Apple releases a fourth beta of its OS} &	\textsc{Neutral} &	  <'bl> \textit{Apple}: \textsc{neutral} \\
			\hline
            <AlmAystrw ... AlAs.twrT rwjer federer mlk Al-l`b alxlfy> 
        	< mn Ajml lq.tAth> 
            
            \textit{The maestro ... the legend Roger Federer king of the backhand game one of his best shots}  & \textsc{HighlyPositive}	& <federer> \textit{Federer}: \textsc{HighlyPositive} \\ \hline
			< Al-lAji'wn ywAjhwn Al.s`wbAt >
            
            \textit{Refugees are facing difficulties}	& \textsc{WeaklyNegative}	& <Al-lAji'wn> \textit{Refugees} : \textsc{Neutral} \\
			\hline
			%Who are you tomorrow? Will you make me smile or just bring me sorrow? \#HottieOfTheWeek Demi Lovato	& \textsc{Neutral}	& Demi Lovato: \textsc{Positive} \\
			%\hline
		\end{tabular}
        }
	\caption{Some Arabic example annotations that we provided to the annotators.}
	\label{T:annotation-arabic-examples}
\end{table*}

\subsection{Annotation using CrowdFlower}

We used CrowdFlower to annotate the new training and testing tweets. The annotators were asked to indicate the overall polarity of the tweet (on a five-point scale) as well as the polarity of the tweet towards the given target topic (again, on a five-point scale), as described in \cite{nakov-EtAl:2016:SemEval1}.
We also provided additional examples, some of which are shown in Tables~\ref{T:annotation-examples} and ~\ref{T:annotation-arabic-examples}. In particular, we stressed that topic-level positive or negative sentiment needed to express an opinion about the topic itself rather than about a positive or a negative event occurring in the context of the topic (see for example, the third row of Table~\ref{T:annotation-arabic-examples}).

Each tweet was annotated by at least five people, and we created many hidden tests for quality control, which we used to reject annotations by contributors who missed a large number of the hidden tests. We also created pilot runs, which helped us adjust the annotation instructions until we found, based on manual inspection, the quality of the annotated tweets to be satisfactory.

%TODO Preslav -- discussion of english annnotations, IAA for english and arabic

\noindent For Arabic, the contributors tended to annotate somewhat conservatively, and thus a very small number of \textsc{HighlyPositive} and \textsc{HighlyNegative} annotations were consolidated, despite us having provided examples of such annotations. 

\subsection{Consolidating the Annotations} 

As the annotations are on a five-point scale, where the
expected agreement is lower, we used a two-step procedure. If three
out of the five annotators agreed on a label, we accepted the label.
Otherwise, we first mapped the categorical labels to the integer
values $-$2, $-$1, 0, 1, 2. Then we calculated the average, and
finally we mapped that average to the closest integer value. In order
to counter-balance the tendency of the average to stay away from the extreme values
$-$2 and 2, and also to prefer 0, we did not use rounding at $\pm$0.5
and $\pm$1.5, but at $\pm$0.4 and $\pm$1.4 instead.
Finally, note that the values $-$2, $-$1, 0, 1, 2 are to be interpreted as \textsc{StronglyNegative}, \textsc{WeaklyNegative}, \textsc{Neutral}, \textsc{WeaklyPositive}, and \textsc{StronglyPositive}, respectively.

\subsection{Data Statistics}

\begin{table*} 
\centering
\begin{tabular}{|l|c|c|c|c|c|c|c|c|}
\hline
 &  &  &  \multicolumn{2}{c|}{\bf{Positive}} & \bf{Neutral}  & \multicolumn{2}{c|}{\bf{Negative}}  &  \\
\bf{Dataset} & \bf{Subtask} & \bf{Topics} & \bf{2} & \bf{1} & \bf{0} & \bf{$-$1} & \bf{$-$2} & Total \\
\hline
\bf{Train} & A & N/A & \multicolumn{2}{c|}{19,902} & 22,591 & \multicolumn{2}{c|}{7,840} & 50,333 \\
 & B, D & 373 & \multicolumn{2}{c|}{14,951} & 1,544 & \multicolumn{2}{c|}{4,013} & 20,508 \\
 & C, E & 200 & 1,020 & 12,922 & 12,993 & 3,398 & 299 & 30,632 \\
 \hline
\bf{Test} & A & N/A & \multicolumn{2}{c|}{2375}  & 5,937 & \multicolumn{2}{c|}{3,972} & 12,284 \\
 & B, D & 125 & \multicolumn{2}{c|}{2,463}  & ---  &  \multicolumn{2}{c|}{3,722} & 6,185 \\
 & C, E & 125 & 131 & 2,332 & 6,194 & 3,545 & 177 & 12,379  \\
 \hline
\end{tabular}
\caption{Statistics about the English training and testing datasets. The training data is the aggregate of all data from prior years, while the testing data is new.}
%A breakdown of prior years can be found in Nakov et al (\citeyear{Nakov:2016ty}).}
\label{T:English_Data}
\end{table*}

\begin{table*} 
\centering
\begin{tabular}{|l|c|c|c|c|c|c|c|c|}
\hline
 &  &  &  \multicolumn{2}{c|}{\bf{Positive}} & \bf{Neutral}  & \multicolumn{2}{c|}{\bf{Negative}}  &  \\
\bf{Dataset} & \bf{Subtask} & \bf{Topics} & \bf{2} & \bf{1} & \bf{0} & \bf{$-$1} & \bf{$-$2} & Total \\
\hline
\bf{Train} & A & N/A & \multicolumn{2}{c|}{743} & 1,470 & \multicolumn{2}{c|}{1,142} & 3,355 \\
 & B, D & 34 & \multicolumn{2}{c|}{885} & --- &  \multicolumn{2}{c|}{771} &  1,656 \\
 & C, E & 34 & 1 & 884 & 1699 & 770 &1  & 3,355  \\
 \hline
\bf{Test} & A & N/A & \multicolumn{2}{c|}{1,514}  & 2,364  & \multicolumn{2}{c|}{2,222} &  6,100 \\
 & B, D & 61  & \multicolumn{2}{c|}{1,561}  & ---  &  \multicolumn{2}{c|}{1,196} & 2,757  \\
 & C, E & 61 & 13 & 1,548 & 3,343  & 1,175 &  21 & 6,100 \\
 \hline
\end{tabular}
\caption{Statistics about the newly collected Arabic training and testing datasets.}
\label{T:Arabic_Data}
\end{table*}

The English training and development data this year consisted of the data from all previous editions of this task ~\cite{Nakov:2013ty,rosenthal-EtAl:2014:SemEval,rosenthal-EtAl:2015:SemEval,Nakov:2016rm}. Unlike in previous years, we did not set aside data to assess progress compared to prior years. Therefore, we allowed all data to be used for training and development.
%giving participants the option to create their own development splits or to use the default splits. 

\noindent For evaluation, we used the newly-created data described in the previous subsection. Tables~\ref{T:English_Data} and ~\ref{T:Arabic_Data} show the statistics for the English and Arabic data. For English, we only show the aggregate statistics for the training data; the breakdown from prior years can be found in \cite{nakov-EtAl:2016:SemEval1}. Note that the same tweets were annotated for multiple subtasks, so there is overlap between the tweets across the tasks. Duplicates may have occurred where the same tweet was extracted for multiple topics.

%To Do: It may be nice to have a breakdown of pos/neg across topic. For example, what is average ratio of sentiment per topic? X topics are mostly positive, Y topics are mostly negative...

As Arabic is a new language this year, we created for it a default train-development split of the Arabic data for the participants to use if they wished to do so. 

\subsection{Data Distribution}

As in previous years, we provided the participants with a script\footnote{\url{https://github.com/seirasto/twitter_download}} to download the training tweets given IDs. In addition, this year we also included in the script the option to download basic user information for the author of each tweet: user id, follower count, status count, description, friend count, location, language, name, and time zone. To ensure a fair evaluation, the test set was provided via download and included the tweets as well as the basic user information provided by the download script. The training and the test data is available for downloaded on our task page.\footnote{\url{http://alt.qcri.org/semeval2017/task4/index.php?id=data-and-tools}}

\section{Evaluation Measures}
\label{sec:eval}

%TODO Preslav: please check that these are correct. 

This section describes the evaluation measures for our five subtasks. Note that for Subtasks B to E, the datasets are each subdivided into a number of topics, and the subtask needs to be carried out independently for each topic. As a
result, each of the evaluation measures will be ``macroaveraged'' across the topics, i.e.,~we compute the measure
individually for each topic, and we then average the results across the topics.

\subsection{Subtask A: Overall Sentiment of a Tweet} 

Our primary measure is AvgRec, or \emph{average recall}, which is recall averaged across the \textsc{Positive} (P), \textsc{Negative} (N), and \textsc{Neutral} (U) classes. This measure has desirable theoretical properties \cite{Sebastiani:2015zl}, and is also the one we use as primarily for Subtask B.
%Given the confusion matrix shown in Table \ref{tab:threewaycontingencytable}
It is computed as follows:

\begin{equation}
  \begin{aligned}
 	  \label{eq:rho}
   AvgRec & = \frac{1}{3}(R^{P}+R^{N}+R^{U}) \\
  \end{aligned}
\end{equation}

\noindent where $R^{P}$, $R^{N}$ and $R^{U}$ refer to recall with respect to the \textsc{Positive}, the \textsc{Negative}, and the \textsc{Neutral} class, respectively.
See \cite{nakov-EtAl:2016:SemEval1} for more detail.

\emph{AvgRec} ranges in $[0,1]$, where a value of 1 is achieved only by the perfect
classifier (i.e., the classifier that correctly classifies all items), a value of 0 is achieved only by the perverse classifier (the classifier that
misclassifies all items), while $0.3333$ is both (\emph{i}) the value for a trivial classifier (i.e.,~one that assigns all tweets
to the same class -- be it \textsc{Positive}, \textsc{Negative}, or \textsc{Neutral}),
and (\emph{ii}) the expected value of a random classifier. 

The advantage of
\emph{AvgRec} over ``standard'' accuracy is that it is more robust to
class imbalance.  The accuracy of the
majority-class classifier is the relative frequency (aka
``prevalence'') of the majority class, that may be much higher than
0.5 if the test set is imbalanced.  Standard $F_{1}$ is also sensitive
to class imbalance for the same reason. Another advantage of
\emph{AvgRec} over $F_{1}$ is that \emph{AvgRec} is invariant with respect
to switching \textsc{Positive} with \textsc{Negative}, while $F_{1}$
is not.  See \cite{Sebastiani:2015zl} for more detail on \emph{AvgRec}.

We further use two secondary measures: accuracy and $F_{1}^{PN}$.
The latter was the primary evaluation measure for Subtask A in previous editions of the task. It is macro-average $F_1$, calculated over the \textsc{Positive} and the \textsc{Negative} classes (note the exclusion of \textsc{Neutral}).
This year, we demoted $F_{1}^{PN}$ to a secondary evaluation measure. It is calculated as follows:

\begin{equation}
  \begin{aligned}
 	  \label{eq:F1}
   F_1^{PN} & = \frac{1}{2}(F_1^{P}+F_1^{N}) \\
  \end{aligned}
\end{equation}

\noindent where $F_1^{P}$ and $F_1^{N}$ refer to $F_1$ with respect to the \textsc{Positive} and the \textsc{Negative} class, respectively.

%where 

%\begin{equation}
%  \begin{aligned}
%    \label{eq:rho_P}
%    \rho^{P} & = \displaystyle\frac{PP}{PP+UP+NP} \\
%     \end{aligned}
%\end{equation}

%\begin{equation}
 % \begin{aligned}
  %  \label{eq:rho_N}
   % \rho^{N} & = \displaystyle\frac{NN}{PN+UN+NN} \\
    % \end{aligned}
%\end{equation}

%\begin{equation}
 % \begin{aligned}
  %  \label{eq:rho_U}
   % \rho^{U} & = \displaystyle\frac{UU}{PU+UU+NU}\\
    % \end{aligned}
%\end{equation}

%\renewcommand{\tabcolsep}{0.1cm} \renewcommand{\arraystretch}{1.30}
%\begin{table}[tbh]
 % \begin{center}
  %  \resizebox{\columnwidth}{!} {
   % \begin{tabular}{|c|c||c|c|c|}
    %  \hline
     % \multicolumn{2}{|c||}{\mbox{}} & \multicolumn{3}{c|}{\bf Gold Standard} \\
     % \cline{3-5}
      %\multicolumn{2}{|c||}{\mbox{}} & \textsc{Positive} & \textsc{Neutral} & \textsc{Negative} \\
      %\hline\hline
      %\multirow{3}{*}{\begin{sideways}\bf Predicted\end{sideways}} & \textsc{Positive} & PP & PU & PN \\
    %  \cline{2-5}
     % & \textsc{Neutral} & UP & UU & UN \\
      %\cline{2-5}
      %& \textsc{Negative} & NP & NU & NN \\
    %  \hline
    %\end{tabular}
  %  }
%\caption{
%\label{tab:threewaycontingencytable} The confusion matrix for Subtask
%A. Cell $XY$ stands for ``the number of tweets that the classifier labeled $X$
%and the gold standard labels as $Y$''. $P$, $U$, $N$ stand for
%\textsc{Positive}, \textsc{Neutral}, \textsc{Negative}, respectively.}
%\end{center}
%\end{table}

\subsection{Subtask B: Topic-Based Classification on a 2-point Scale} 

As in 2016, our primary evaluation measure for subtask B is average recall, or AvgRec
%, which is averaged over the \textsc{Positive} and the \textsc{Negative} class 
(note that there are only two classes for this subtask): 

\begin{equation}
  \begin{aligned}
    \label{eq:rhoPN}
    AvgRec & = \frac{1}{2}(R^{P}+R^{N}) \\
  \end{aligned}
\end{equation}

We further use accuracy and $F_1$ as secondary measures for subtask B.
Finally, as subtask B is topic-based, we computed each metric individually for each topic, and we then averaged the result across the topics to yield the final score.

\subsection{Subtask C: Topic-based Classification on a 5-point Scale} 

\noindent Subtask C is an \emph{ordinal classification} (also
known as \emph{ordinal regression}) task, in which each tweet must be
classified into exactly one of the classes in
$\mathcal{C}$=\{\textsc{HighlyPositive}, \textsc{Positive},
\textsc{Neutral}, \textsc{Negative}, \textsc{HighlyNegative}\},
represented in our dataset by numbers in \{$+2$,$+1$,$0$,$-1$,$-2$\}, with a
total order defined on $\mathcal{C}$.

We adopt an evaluation measure that takes the order of the five classes into account. For instance, misclassifying a \textsc{HighlyNegative} example as \textsc{HighlyPositive} is a bigger mistake than misclassifying it as \textsc{Negative} or as \textsc{Neutral}.

%\begin{equation}
%\begin{aligned}
 % \label{eq:macroMAE}MAE^{M}(h,Te) =
  %\frac{1}{|\mathcal{C}|}\sum_{j=1}^{|\mathcal{C}|}\frac{1}{|Te_{j}|}\sum_{\mathbf{x}_{i}\in
  %Te_{j}}|h(\mathbf{x}_{i})-y_{i}| \nonumber
%  \end{aligned}
%\end{equation}
%

%\noindent where $y_{i}$ denotes the true label of item
%$\mathbf{x}_{i}$, $h(\mathbf{x}_{i})$ is its predicted label,
%$Te_{j}$ denotes the set of test documents whose true class is
%$c_{j}$, $|h(\mathbf{x}_{i})-y_{i}|$ denotes the ``distance'' between
%classes $h(\mathbf{x}_{i})$ and $y_{i}$ (e.g., the distance between
%\textsc{HighlyPositive} and \textsc{Negative} is 3), and the ``M''
%superscript indicates ``macroaveraging''.

As in SemEval-2016 Task 4, we use \emph{macro-average mean absolute
error} ($MAE^{M}$) as the main ordinal classification measure:
%It is defined as follows:
%
\begin{equation}
  \label{eq:macroMAE}MAE^{M}(h,Te) =
  \frac{1}{|\mathcal{C}|}\sum_{j=1}^{|\mathcal{C}|}\frac{1}{|Te_{j}|}\sum_{\mathbf{x}_{i}\in
  Te_{j}}|h(\mathbf{x}_{i})-y_{i}| \nonumber
\end{equation}
\noindent where $y_{i}$ denotes the true label of item
$\mathbf{x}_{i}$, $h(\mathbf{x}_{i})$ is its predicted label,
$Te_{j}$ denotes the set of test documents whose true class is
$c_{j}$, $|h(\mathbf{x}_{i})-y_{i}|$ denotes the ``distance'' between
classes $h(\mathbf{x}_{i})$ and $y_{i}$ (e.g., the distance between
\textsc{HighlyPositive} and \textsc{Negative} is 3), and the ``M''
superscript indicates ``macroaveraging''.

The advantage of $MAE^{M}$ over ``standard'' mean absolute error,
which is defined as
\begin{equation}\label{eq:microMAE}MAE^{\mu}(h,Te)=\frac{1}{
  |Te|}\sum_{\mathbf{x}_{i}\in
  Te}|h(\mathbf{x}_{i})-y_{i}|\end{equation}
\noindent is that it is robust to class imbalance (which is
useful, given the imbalanced nature of our dataset). On perfectly
balanced datasets $MAE^{M}$ and $MAE^{\mu}$ are equivalent.

\noindent $MAE^{M}$ is an extension of macro-average recall for ordinal regression; yet, it is a measure of error, and thus lower values are better. 
We also use $MAE^{\mu}$ as a secondary measure, in order to provide better consistency with Subtasks A and B. These measures are computed for each
topic, and the results are then averaged across all topics to yield
the final score. 
See \cite{Baccianella:2009qd} for more detail about $MAE^{M}$ and $MAE^{\mu}$.

\subsection{Subtask D: Tweet Quantification on a 2-point Scale}

Subtask D assumes a binary quantification setup, in which each tweet is classified as \textsc{Positive} or \textsc{Negative}, and the distribution across classes must be estimated. %The task is to
%compute an estimate $\hat{p}(c_{j})$ of
% the true distribution $p$ of the test tweets across the classes in
% $\mathcal{C}$ (i.e., the task is to compute
%the relative frequency (in the test set) of each of the classes. 
The difference with binary classification is that errors of different polarity (e.g., a false positive and a false negative for the same
class) can compensate for each other in quantification. Quantification is thus
a more lenient task than classification,
since a perfect classifier is also a perfect quantifier,
but a perfect quantifier is not necessarily a perfect classifier.

For evaluating binary quantification, we keep the \emph{Kullback-Leibler Divergence} \textit{(KLD)} measure used in 2016 along with additive smoothing \cite{nakov-EtAl:2016:SemEval1,Forman:2005fk}. 
KLD was proposed as a quantification measure in \cite{Forman:2005fk}, and is
defined as follows:
\begin{equation}
  \label{eq:KLD}
  KLD(\hat{p},p,\mathcal{C}) = \sum_{c_{j}\in \mathcal{C}} 
  p(c_{j})\log_e\frac{p(c_{j})}{\hat{p}(c_{j})}
\end{equation}

$KLD$ is a measure of the error made in estimating a true
distribution $p$ over a set $\mathcal{C}$ of classes by means of a
predicted distribution $\hat{p}$. Like $MAE^{M}$, $KLD$ is a measure of error, which means that lower values are
better. $KLD$ ranges between 0 (best) and $+\infty$ (worst).

Note that the upper bound of $KLD$ is not finite since Equation
\ref{eq:KLD} has predicted prevalences, and not true prevalences, at
the denominator: that is, by making a predicted prevalence
$\hat{p}(c_{j})$ infinitely small we can make $KLD$ infinitely large.
% $KLD$ may be undefined in some cases. While the case in which
% $p(c_{j})=0$ is not problematic (since continuity arguments indicate
% that $0 \log \frac{0}{a}$ should be taken to be 0 for any $a\geq
% 0$), the case in which $\hat{p}(c_{j})=0$ and $p(c_{j})>0$ is indeed
% problematic, since $a\log\frac{a}{0}$ is undefined for $a>0$.
To solve this problem, in computing $KLD$ we smooth both $p(c_{j})$
and $\hat{p}(c_{j})$ via additive smoothing, i.e.,
\begin{equation}
  \begin{aligned}
    \label{eq:smoothing}
    p^{s}(c_{j})= &
    \frac{p(c_{j})+\epsilon}{(\displaystyle\sum_{c_{j}\in
    \mathcal{C}}p(c_{j}))+\epsilon\cdot|\mathcal{C}|} \\ = &
    \frac{p(c_{j})+\epsilon}{1+\epsilon\cdot|\mathcal{C}|}
  \end{aligned}
\end{equation}

\noindent where $p^{s}(c_{j})$ denotes the smoothed version of
$p(c_{j})$ and the denominator is just a normalizer (same for
the $\hat{p}^{s}(c_{j})$'s); the quantity $\epsilon=\frac{1}{2\cdot
|Te|}$ is used as a smoothing factor, where $Te$ denotes the test dataset.

The smoothed versions of $p(c_{j})$ and $\hat{p}(c_{j})$ are used
in place of their original versions in Equation \ref{eq:KLD}; as a
result, $KLD$ is always defined and still returns a value of 0 when
$p$ and $\hat{p}$ coincide.

Like $MAE^{M}$, $KLD$ is a measure of error, which means that lower values are better. We further use two secondary error-based evaluation measures: \emph{absolute error} (AE), and \emph{relative absolute error} (RAE).

Again, the measures are computed individually for each topic, and the results are averaged across the topics to yield the final score.

\subsection{Subtask E: Tweet Quantification on a 5-point Scale} 

Subtask E is an ordinal quantification task. As in binary quantification, the goal is to compute the distribution across classes, this time assuming a quantification setup.
%the task is to compute an estimate $\hat{p}(c_{j})$ of the relative
%frequency $p(c_{j})$ in the test tweets of all the classes $c_{j}\in
%\mathcal{C}$.

Here each tweet belongs exactly to one of the classes in
$\mathcal{C}$=\{\textsc{HighlyPositive}, \textsc{Positive},
\textsc{Neutral}, \textsc{Negative}, \textsc{HighlyNegative}\}, where
there is a total order on $\mathcal{C}$. As in binary quantification,
the task is to compute an estimate $\hat{p}(c_{j})$ of the relative
frequency $p(c_{j})$ in the test tweets of all the classes $c_{j}\in
\mathcal{C}$.

The measure we adopt for ordinal quantification is the \emph{Earth Mover's Distance}
\cite{Rubner:2000fk}, also known as the \emph{Vaser\u{s}te\u{\i}n
metric} \cite{Ruschendorf:2001mz}, a measure well-known in the field
of computer vision. $EMD$ is currently the only known measure for
ordinal quantification. It is defined for the general case in
which a distance $d(c',c'')$ is defined for each
$c',c''\in\mathcal{C}$. When there is a total order on the classes in
$\mathcal{C}$ and $d(c_i,c_{i+1})=1$ for all
$i\in\{1,...,(\mathcal{C}-1)\}$, the Earth Mover's Distance is defined as
\begin{equation}
  \label{eq:EMD}
  EMD(\hat{p},p) = \sum_{j=1}^{|\mathcal{C}|-1}|\sum_{i=1}^{j}\hat{p}(c_{i})-\sum_{i=1}^{j}p(c_{i})|
\end{equation}
\noindent and can be computed in $|\mathcal{C}|$ steps from the
estimated and true class prevalences.

Like $KLD$, $EMD$ is a measure of error, so lower values are better; $EMD$ ranges between 0 (best) and
$|\mathcal{C}|-1$ (worst). See \cite{Esuli:2010fk} for more detail on $EMD$.

\noindent As before, $EMD$ is computed individually for each topic, and the
results are then averaged across all topics to yield the final score.
For more detail on $EMD$, the reader is referred to \cite{Esuli:2010fk} and to last year's task description paper \cite{nakov-EtAl:2016:SemEval1}.

\section{Participants and Results}
\label{sec:results}

A total of 48 teams participated in SemEval-2017 Task 4 this year. As in previous years, the most popular subtask this year was Subtask A, with 38 teams participating in the English subtask A, and 8 teams participating in the Arabic subtask A. Overall, there were 46 teams who participated in some English subtask and 9 teams that participated in some Arabic subtask.
%Many teams ( ) signed up for all the English versions of the task (Subtasks A-E) but not for the Arabic version;
%Z teams participated in Subtasks A for both English and Arabic, and W teams participated only in Arabic subtasks.
There were 28 teams that participated in a subtask other than subtask A. Moreover, two teams (OMAM and ELiRF-UPV) 
participated in all English and in all Arabic subtasks. There were 9 teams that participated in the topic versions of the subtasks but not in subtask A, reflecting a growing interest among researchers in developing systems for topic-specific analysis. 
%(NB: I AM basing these counts on the people who responded, there might be more.)

\subsection{Common Resources and Methods}

In terms of methods, the use of deep learning stands out in particular, and we also see an increase over the last year.
There were at least 20 teams 
%who responded to our survey
who used deep learning and neural network methods such as CNN and LSTM networks. %In particular, deep learning tools such as Keras, TensorFlow and DL4J were commonly used. 
%Almost as popular was SVM (18 teams) and Liblinear (18+3=21)
Supervised SVM and Liblinear were also very popular, with several participants combining SVM with neural network methods or SVM with dense word embedding features. Other teams used classifiers such as Maximum Entropy, Logistic Regression, Random Forest, Na\"{i}ve Bayes classifier, and Conditional Random Fields.    %especially with the increasing popularity of deep learning tools like Keras and NLP4J.
%Only one team (INGEOTEC, who participated in both languages for Subtask A) cited having employed a cross-lingual (multilingual) method. They ranked 7th in the English subtask and 4th in the Arabic subtask. 
%In this section we describe the common trends in resources and methods across subtasks to provide a glimpse of useful tools available. 

Common software used included Python (with the sklearn and numpy libraries), Java, Tensorflow, Weka, NLTK, Keras, Theano, and Stanford CoreNLP. The most common external datasets used were sentiment140 as a lexicon, pre-trained word2vec embeddings. Many teams further gathered additional tweets using the Twitter API that were not annotated for sentiment. These were used for distant supervision, lexicon building, and word vector training.
%Deep learning including RNN, CNN, and LSTM, and word embeddings continue to be common approaches in addition to more classical supervised approaches of SVM, Logistic Regression, Random Forest, and Naive Bayes.

In the following subsections, we present the results and the ranking for each subtask, and we highlight the best-performing systems for each subtask.

\begin{table}[h!]
\centering
\begin{small}
\renewcommand{\arraystretch}{1.0}% Tighter
\begin{tabular}{|c|l|l|l|l|}
\hline
      \bf \# & \multicolumn{1}{c|}{\bf System} & \multicolumn{1}{c|}{\it AvgRec} & \multicolumn{1}{c|}{$F_{1}^{PN}$} & \multicolumn{1}{c|}{$Acc$} \\
\hline
\bf 1 & DataStories & \bf 0.681$_{1}$ & 0.677$_{2}$ & 0.651$_{5}$ \\
& BB\_twtr & \bf 0.681$_{1}$ & 0.685$_{1}$ & 0.658$_{3}$ \\
\bf 3 & LIA & \bf 0.676$_{3}$ & 0.674$_{3}$ & 0.661$_{2}$ \\
\bf 4 & Senti17 & \bf 0.674$_{4}$ & 0.665$_{4}$ & 0.652$_{4}$ \\
\bf 5 & NNEMBs & \bf 0.669$_{5}$ & 0.658$_{5}$ & 0.664$_{1}$ \\
\bf 6 & Tweester & \bf 0.659$_{6}$ & 0.648$_{6}$ & 0.648$_{6}$ \\
\bf 7 & INGEOTEC & \bf 0.649$_{7}$ & 0.645$_{7}$ & 0.633$_{11}$ \\
\bf 8 & SiTAKA & \bf 0.645$_{8}$ & 0.628$_{9}$ & 0.643$_{9}$ \\
\bf 9 & TSA-INF & \bf 0.643$_{9}$ & 0.620$_{11}$ & 0.616$_{17}$ \\
\bf 10 & UCSC-NLP & \bf 0.642$_{10}$ & 0.624$_{10}$ & 0.565$_{30}$ \\
\bf 11 & HLP@UPENN & \bf 0.637$_{11}$ & 0.632$_{8}$ & 0.646$_{8}$ \\
\bf 12 & YNU-HPCC & \bf 0.633$_{12}$ & 0.612$_{15}$ & 0.647$_{7}$ \\
& SentiME++ & \bf 0.633$_{12}$ & 0.613$_{13}$ & 0.601$_{23}$ \\
\bf 14 & ELiRF-UPV & \bf 0.632$_{14}$ & 0.619$_{12}$ & 0.599$_{24}$ \\
\bf 15 & ECNU & \bf 0.628$_{15}$ & 0.613$_{13}$ & 0.630$_{12}$ \\
\bf 16 & TakeLab & \bf 0.627$_{16}$ & 0.607$_{16}$ & 0.628$_{14}$ \\
\bf 17 & DUTH & \bf 0.621$_{17}$ & 0.605$_{17}$ & 0.640$_{10}$ \\
\bf 18 & CrystalNest & \bf 0.619$_{18}$ & 0.593$_{19}$ & 0.629$_{13}$ \\
\bf 19 & deepSA & \bf 0.618$_{19}$ & 0.587$_{20}$ & 0.616$_{17}$ \\
\bf 20 & NILC-USP & \bf 0.612$_{20}$ & 0.595$_{18}$ & 0.617$_{16}$ \\
\bf 21 & Ti-Senti & \bf 0.607$_{21}$ & 0.577$_{22}$ & 0.627$_{15}$ \\
\bf 22 & BUSEM & \bf 0.605$_{22}$ & 0.587$_{20}$ & 0.603$_{22}$ \\
\bf 23 & EICA & \bf 0.595$_{23}$ & 0.555$_{24}$ & 0.599$_{24}$ \\
\bf 24 & OMAM & \bf 0.590$_{24}$ & 0.542$_{26}$ & 0.615$_{19}$ \\
\bf 25 & Adullam & \bf 0.589$_{25}$ & 0.552$_{25}$ & 0.614$_{20}$ \\
\bf 26 & NileTMRG & \bf 0.578$_{26}$ & 0.515$_{32}$ & 0.606$_{21}$ \\
\bf 27 & Amobee-C-137 & \bf 0.575$_{27}$ & 0.520$_{30}$ & 0.587$_{27}$ \\
\bf 28 & ej-za-2017 & \bf 0.571$_{28}$ & 0.539$_{27}$ & 0.582$_{28}$ \\
& LSIS & \bf 0.571$_{28}$ & 0.561$_{23}$ & 0.521$_{34}$ \\
\bf 30 & XJSA & \bf 0.556$_{30}$ & 0.519$_{31}$ & 0.575$_{29}$ \\
\bf 31 & Neverland-THU & \bf 0.555$_{31}$ & 0.507$_{33}$ & 0.597$_{26}$ \\
\bf 32 & MI\&T-Lab & \bf 0.551$_{32}$ & 0.522$_{29}$ & 0.561$_{31}$ \\
\bf 33 & diegoref & \bf 0.546$_{33}$ & 0.527$_{28}$ & 0.540$_{33}$ \\
\bf 34 & xiwu & \bf 0.479$_{34}$ & 0.365$_{34}$ & 0.547$_{32}$ \\
\bf 35 & SSN\_MLRG1 & \bf 0.431$_{35}$ & 0.344$_{35}$ & 0.439$_{35}$ \\
\bf 36 & YNUDLG & \bf 0.340$_{36}$ & 0.201$_{37}$ & 0.387$_{36}$ \\
\bf 37 & WarwickDCS & \bf 0.335$_{37}$ & 0.221$_{36}$ & 0.382$_{37}$ \\
& Avid & \bf 0.335$_{37}$ & 0.163$_{38}$ & 0.206$_{38}$ \\
\hline
B1 & All \textsc{Positive} & \bf 0.333 & 0.162 & 0.193\\
B2 & All \textsc{Negative} & \bf 0.333 & \bf 0.244 & 0.323\\
B3 & All \textsc{Neutral}  & \bf 0.333 & 0.000 & \bf 0.483\\
\hline
\end{tabular}
\caption{\textbf{Results for Subtask A ``Message Polarity Classification'', English.} 
The systems are ordered by average recall \emph{AvgRec} (higher is better). In each column, the rankings according to the corresponding measure are indicated with a subscript. B$x$ indicates a baseline.}
\label{table:ResultsSubtaskA:English}
\end{small}
\end{table}

\begin{table}[h!]
\centering
\begin{small}
\renewcommand{\arraystretch}{1.0}% Tighter
\begin{tabular}{|c|l|l|l|l|}
\hline
      \bf \# & \multicolumn{1}{c|}{\bf System} & \multicolumn{1}{c|}{\it AvgRec} & \multicolumn{1}{c|}{$F_{1}^{PN}$} & \multicolumn{1}{c|}{$Acc$} \\
\hline
\bf 1 & NileTMRG & \bf 0.583$_{1}$ & 0.610$_{1}$ & 0.581$_{1}$ \\
\bf 2 & SiTAKA & \bf 0.550$_{2}$ & 0.571$_{2}$ & 0.563$_{2}$ \\
\bf 3 & ELiRF-UPV & \bf 0.478$_{3}$ & 0.467$_{4}$ & 0.508$_{3}$ \\
\bf 4 & INGEOTEC & \bf 0.477$_{4}$ & 0.455$_{5}$ & 0.499$_{4}$ \\
\bf 5 & OMAM & \bf 0.438$_{5}$ & 0.422$_{6}$ & 0.430$_{8}$ \\
& LSIS & \bf 0.438$_{5}$ & 0.469$_{3}$ & 0.445$_{6}$ \\
\bf 7 & Tw-StAR & \bf 0.431$_{7}$ & 0.416$_{7}$ & 0.454$_{5}$ \\
\bf 8 & HLP@UPENN & \bf 0.415$_{8}$ & 0.320$_{8}$ & 0.443$_{7}$ \\
\hline
B1 & All \textsc{Positive} & \bf 0.333 & 0.199 & 0.248\\
B2 & All \textsc{Negative} & \bf 0.333 & \bf 0.267 & 0.364\\
B3 & All \textsc{Neutral}  & \bf 0.333 & 0.000 & \bf 0.388\\
\hline
\end{tabular}
\caption{\textbf{Results for Subtask A ``Message Polarity Classification'', Arabic.} 
The systems are ordered by average recall \emph{AvgRec} (higher is better). In each column, the rankings according to the corresponding measure are indicated with a subscript. B$x$ indicates a baseline.}
\label{table:ResultsSubtaskA:Arabic}
\end{small}
\end{table}

\subsection{Results for Subtask A: Overall Sentiment in a Tweet}

Tables \ref{table:ResultsSubtaskA:English} and \ref{table:ResultsSubtaskA:Arabic} show the results for Subtask A in English and Arabic, respectively, where the teams are ranked by macro-average recall.
%and also scored using $F_1^{PN}$ and accuracy. 

\paragraph{For English} the best ranking teams were \emph{BB\_twtr} and \emph{DataStories}, both achieving a macro-average recall of 0.681.
% which is better by .005 absolute than the score for \emph{LIA}, the next best team.  
Both top teams used deep learning;  \emph{BB\_twtr} used an ensemble of LSTMs and CNNs with multiple convolution operations, while \emph{DataStories} used deep LSTM networks with an attention mechanism.

 %with no external training datasets. 
% but the word vectors were trained on millions of external tweets.
%trained on top of GloVe word embeddings.
%and initialized using a collection of 330 million tweets that they collected.
 %They reported using 100 million unlabeled English tweets, possibly also for training word vectors.
Both teams participated in all English subtasks and were also ranked in first (\emph{BB\_twtr}) and second (\emph{DataStories}) place for subtasks B-D; \emph{BB\_twtr} was also ranked first for subtask E.

The top 5 teams for English were very closely scored. The following four best-ranked teams all used deep learning or deep learning ensembles. %CNNs, or CNN ensembles.
Three of the top-10 scoring teams (\emph{INGEOTEC}, \emph{SiTAKA}, and \emph{UCSC-NLP}) used SVM classifiers instead, with various surface, lexical, semantic, and dense word embedding features. The use of ensembles clearly stood out, with five of the top-10 scoring systems (\emph{BB\_twtr}, \emph{LIA}, \emph{NNEMBs}, \emph{Tweester}, and \emph{INGEOTEC}) using ensembles, hybrid, stacking or some kind of mix of learning methods. All teams beat the baseline on macro-average recall; however, a few teams did not beat the harsher average F-measure and accuracy baselines.

\paragraph{For Arabic} the best-ranked team was \emph{NileTMRG}, and it achieved a score of 0.583. They used a Na\"{i}ve Bayes classifier with a combination of lexical and sentiment features; they further augmented the training dataset to about 13K examples using external tweets. 
%as well as a weighted sentiment lexicon. %Their score was very similar for the English subtask, where they ranked 26th.
%They ranked 26th for the English task, where they used a deep learning system.
The \emph{SiTAKA} team was ranked second with a score of 0.55. Their system used a feature-rich SVM with lexical features and embedding representations.
%with different pooling functions. 
%such as negated-bag-of-words, and embedding representations with different pooling functions. 
%They used a rich feature SVM with bag of words, lexicon and embedding features. They came 8th in English, where they scored 0.645. 
%They presumably used the same system as they did for English (where they scored 0.645), a rich feature SVM with bag of negated words (BonW) in addition to word embedding features and a set of lexicons.
%The remaining teams, with the exception of 1w-StAr, also participated in English, with similar drops in their respective scores when moving to Arabic.
Except for \emph{EliRF-UPV}, who used multi-layer neural networks (CRNNs), the remaining teams used SVM and Na\"{i}ve Bayes classifiers, genetic algorithms, or conditional random fields (CRFs). All teams managed to beat all baselines for all metrics.

\noindent The difference in the absolute scores for the two languages is probably partially due to the difference in the amount of training data available for Arabic, which was much smaller compared English, even when external datasets were taken into account. 
%(We note that the same annotation process and instructions were applied for annotating tweets in both languages).
The results also reflect the linguistic complexity of Arabic as it is used in social media, which is characterized by the abundant use of dialectal forms and spelling variants. Overall, participants preferred to focus on developing Arabic-specific systems (varying in the extent to which they applied Arabic-specific preprocessing) rather than trying to leverage cross-language models that would enable them to use English data to augment their Arabic models.
%We were hoping that some teams would leverage training data in English to augment their Arabic models, but it does not appear that any of them took advantage of the English data. 

%In our system we have defined a new polarity measure based on a set of lexicons and we have added Bag of negated Words (BonW) features in addition to a basic set of features (i.e. BoW, POS,.. etc) and an bedimming based features.
%combined with features described in (El-Beltagy, et al, 2016) and a weighted sentiment lexicon.
%They used the same(?) system for English and Arabic.
%For task A, we employed a Complement Naive Bayes classifier combined with features described in (El-Beltagy, et al, 2016) ) cited above as well as a weighted version on the NU Lexicon described in (El-Beltagy, Samhaa R. 2017. "WeightedNileULex: A Scored Arabic Sentiment Lexicon for Improved Sentiment Analysis", to appear in Book Series on Language Processing, Pattern Recognition and Intelligent Systems: Special Issue on Computational Linguistics, Speech& Image Processing for Arabic Language, Publisher: World Scientific Publishing Co, Editors: Neamat El Gayar, Ching Suen) 
%We developed 2 different deep-learning networks (one model for taskA and another for tasks B,C,D,E). Our systems use Long Short-Term Memory (LSTM) networks augmented with attention mechanism, trained on top of GloVe word embeddings pre-trained on a big collection of Twitter messages (that we collected).

%\newpage
\begin{table}[h!]
\centering
\begin{small}
\renewcommand{\arraystretch}{1.0}% Tighter
\begin{tabular}{|c|l|l|l|l|}
\hline
      \bf \# & \bf System & \multicolumn{1}{c|}{\it AvgRec} & \multicolumn{1}{c|}{$F_1$} & \multicolumn{1}{c|}{$Acc$} \\
\hline
\bf 1 & BB\_twtr & \bf 0.882$_{1}$ & 0.890$_{1}$ & 0.897$_{1}$ \\
\bf 2 & DataStories & \bf 0.856$_{2}$ & 0.861$_{2}$ & 0.869$_{2}$ \\
\bf 3 & Tweester & \bf 0.854$_{3}$ & 0.856$_{3}$ & 0.863$_{3}$ \\
\bf 4 & TopicThunder & \bf 0.846$_{4}$ & 0.847$_{4}$ & 0.854$_{4}$ \\
\bf 5 & TakeLab & \bf 0.845$_{5}$ & 0.836$_{5}$ & 0.840$_{6}$ \\
\bf 6 & funSentiment & \bf 0.834$_{6}$ & 0.824$_{8}$ & 0.827$_{8}$ \\
& YNU-HPCC & \bf 0.834$_{6}$ & 0.816$_{10}$ & 0.818$_{10}$ \\
\bf 8 & WarwickDCS & \bf 0.829$_{8}$ & 0.834$_{6}$ & 0.843$_{5}$ \\
\bf 9 & CrystalNest & \bf 0.827$_{9}$ & 0.822$_{9}$ & 0.827$_{8}$ \\
\bf 10 & Ti-Senti & \bf 0.826$_{10}$ & 0.830$_{7}$ & 0.838$_{7}$ \\
\bf 11 & Amobee-C-137 & \bf 0.822$_{11}$ & 0.801$_{12}$ & 0.802$_{12}$ \\
\bf 12 & SINAI & \bf 0.818$_{12}$ & 0.806$_{11}$ & 0.809$_{11}$ \\
\bf 13 & NRU-HSE & \bf 0.798$_{13}$ & 0.787$_{13}$ & 0.790$_{13}$ \\
\bf 14 & EICA & \bf 0.790$_{14}$ & 0.775$_{14}$ & 0.777$_{16}$ \\
\bf 15 & OMAM & \bf 0.779$_{15}$ & 0.762$_{17}$ & 0.764$_{17}$ \\
\bf 16 & NileTMRG & \bf 0.769$_{16}$ & 0.774$_{15}$ & 0.789$_{15}$ \\
\bf 17 & ELiRF-UPV & \bf 0.766$_{17}$ & 0.773$_{16}$ & 0.790$_{13}$ \\
\bf 18 & DUTH & \bf 0.663$_{18}$ & 0.600$_{18}$ & 0.607$_{18}$ \\
\bf 19 & ej-za-2017 & \bf 0.594$_{19}$ & 0.486$_{21}$ & 0.518$_{19}$ \\
\bf 20 & SSN\_MLRG1 & \bf 0.586$_{20}$ & 0.494$_{20}$ & 0.518$_{19}$ \\
\bf 21 & YNUDLG & \bf 0.516$_{21}$ & 0.499$_{19}$ & 0.499$_{21}$ \\
\bf 22 & TM-Gist & \bf 0.499$_{22}$ & 0.428$_{22}$ & 0.444$_{22}$ \\
\bf 23 & SSK\_JNTUH & \bf 0.483$_{23}$ & 0.372$_{23}$ & 0.412$_{23}$ \\
\hline
B1 & All \textsc{Positive} & \bf 0.500 & 0.285 & 0.398\\
B2 & All \textsc{Negative} & \bf 0.500 & \bf 0.376 & \bf 0.602\\
\hline
\end{tabular}
\caption{\textbf{Results for Subtask B ``Tweet classification according to a two-point scale'', English.} The systems are ordered by average recall \emph{AvgRec} (higher is better). B$x$ indicates a baseline.}
\label{table:ResultsSubtaskB:English}
\end{small}
\end{table}

\begin{table}[h!]
\centering
\begin{small}
\renewcommand{\arraystretch}{1.0}% Tighter
\begin{tabular}{|c|l|l|l|l|}
\hline
      \bf \# & \bf System & \multicolumn{1}{c|}{\it AvgRec} & \multicolumn{1}{c|}{$F_1$} & \multicolumn{1}{c|}{$Acc$} \\
\hline
\bf 1 & NileTMRG & \bf 0.768$_{1}$ & 0.767$_{1}$ & 0.770$_{1}$ \\
\bf 2 & ELiRF-UPV & \bf 0.721$_{2}$ & 0.724$_{2}$ & 0.734$_{2}$ \\
\bf 3 & ASA & \bf 0.693$_{3}$ & 0.670$_{4}$ & 0.672$_{4}$ \\
\bf 4 & OMAM & \bf 0.687$_{4}$ & 0.678$_{3}$ & 0.679$_{3}$ \\
\hline
B1 & All \textsc{Positive} & \bf 0.500 & \bf 0.362 & \bf 0.566\\
B2 & All \textsc{Negative} & 0.500 & 0.303 & 0.434\\
\hline
\end{tabular}
\caption{\textbf{Results for Subtask B ``Tweet classification according to a two-point scale'', Arabic.} The systems are ordered by average recall \emph{AvgRec} (higher is better). B$x$ indicates a baseline.}
\label{table:ResultsSubtaskB:Arabic}
\end{small}
\end{table}

\subsection{Results for Subtasks B and C: Topic-Based Classification}

The results of Subtasks B and C are shown in Tables \ref{table:ResultsSubtaskB:English}--\ref{table:ResultsSubtaskC:Arabic}. 
We can see that the system scores for subtask B are higher than those for subtask A, with the best team achieving 0.882 accuracy for English (compared to 0.681 for subtask A) and 0.768 for Arabic (compared to 0.583 for subtask A). However, this is primarily due to the fact there are two classes for subtask B, while there are three classes for subtask A.

%\newpage
\begin{table}[h!]
\centering
\begin{small}
\renewcommand{\arraystretch}{1.0}% Tighter
\begin{tabular}{|c|l|l|l|}
\hline
      \bf \# & \bf System & \multicolumn{1}{c|}{$MAE^M$} & \multicolumn{1}{c|}{$MAE^{\mu}$} \\
\hline
\bf 1 & BB\_twtr & \bf 0.481$_{1}$ & 0.554$_{6}$ \\
\bf 2 & DataStories & \bf 0.555$_{2}$ & 0.543$_{4}$ \\
\bf 3 & Amobee-C-137 & \bf 0.599$_{3}$ & 0.582$_{10}$ \\
\bf 4 & Tweester & \bf 0.623$_{4}$ & 0.734$_{13}$ \\
\bf 5 & TwiSe & \bf 0.640$_{5}$ & 0.616$_{12}$ \\
\bf 6 & CrystalNest & \bf 0.698$_{6}$ & 0.571$_{9}$ \\
\bf 7 & ELiRF-UPV & \bf 0.806$_{7}$ & 0.586$_{11}$ \\
\bf 8 & EICA & \bf 0.823$_{8}$ & 0.509$_{2}$ \\
\bf 9 & funSentiment & \bf 0.842$_{9}$ & 0.530$_{3}$ \\
\bf 10 & DUTH & \bf 0.895$_{10}$ & 0.544$_{5}$ \\
& OMAM & \bf 0.895$_{10}$ & 0.475$_{1}$ \\
\bf 12 & YNU-HPCC & \bf 0.925$_{12}$ & 0.567$_{8}$ \\
\bf 13 & NRU-HSE & \bf 0.928$_{13}$ & 0.557$_{7}$ \\
\bf 14 & YNU-1510 & \bf 1.262$_{14}$ & 0.764$_{14}$ \\
\bf 15 & SSN\_MLRG1 & \bf 1.325$_{15}$ & 0.985$_{15}$ \\
\hline
B1 &  \textsc{HighlyNegative} & 2.000 & 1.895\\
B2 & \textsc{Negative} & 1.400 & 0.923\\
B3 & \textsc{Neutral} & \bf 1.200 & \bf 0.525\\
B4 & \textsc{Positive} & 1.400 & 1.127\\
B5 & \textsc{HighlyPositive} & 2.000 & 2.105\\
\hline
\end{tabular}
\caption{\textbf{Results for Subtask C ``Tweet classification according to a five-point scale'', English.} The systems are ordered by their $MAE^{M}$ score (lower is better). B$x$ indicates a baseline.}
\label{table:ResultsSubtaskC:English}
\end{small}
\end{table}

\begin{table}[h!]
\centering
\begin{small}
\renewcommand{\arraystretch}{1.0}% Tighter
\begin{tabular}{|c|l|l|l|}
\hline
      \bf \# & \bf System & \multicolumn{1}{c|}{$MAE^M$} & \multicolumn{1}{c|}{$MAE^{\mu}$} \\
\hline
\bf 1 & OMAM & \bf 0.943$_{1}$ & 0.646$_{1}$ \\
\bf 2 & ELiRF-UPV & \bf 1.264$_{2}$ & 0.787$_{2}$ \\
\hline
B1 & \textsc{HighlyNegative} & 2.000 & 2.059\\
B2 & \textsc{Negative} & 1.400 & 1.065\\
B3 & \textsc{Neutral} & \bf 1.200 & \bf 0.458\\
B4 & \textsc{Positive} & 1.400 & 0.946\\
B5 & \textsc{HighlyPositive} & 2.000 & 1.941\\
\hline
\end{tabular}
\caption{\textbf{Results for Subtask C ``Tweet classification according to a five-point scale'', Arabic.} The systems are ordered by their $MAE^{M}$ score (lower is better). B$x$ indicates a baseline.}
\label{table:ResultsSubtaskC:Arabic}
\end{small}
\end{table}

\paragraph{For English} the \emph{BB\_twtr} system, ranked first, modeled topics by concatenating the topical information at the word level.
The second-best system, \emph{DataStories}, also accounted for topics by producing topic annotations and a context-aware attention mechanism. 

%\newpage
\begin{table}[tbh]
\centering
\begin{small}
\renewcommand{\arraystretch}{1.0}% Tighter
\setlength{\tabcolsep}{2pt}
\begin{tabular}{|c|l|l|l|l|}
\hline
      \bf \# & \bf System & \multicolumn{1}{c|}{$KLD$} & \multicolumn{1}{c|}{$AE$} &\multicolumn{1}{c|}{$RAE$} \\
\hline
\bf 1 & BB\_twtr & \bf 0.036$_{1}$ & 0.080$_{1}$ & 0.598$_{1}$ \\
\bf 2 & DataStories & \bf 0.048$_{2}$ & 0.095$_{2}$ & 0.848$_{2}$ \\
\bf 3 & TakeLab & \bf 0.050$_{3}$ & 0.096$_{3}$ & 1.057$_{5}$ \\
\bf 4 & CrystalNest & \bf 0.056$_{4}$ & 0.104$_{5}$ & 1.202$_{6}$ \\
\bf 5 & Tweester & \bf 0.057$_{5}$ & 0.103$_{4}$ & 1.051$_{4}$ \\
\bf 6 & funSentiment & \bf 0.060$_{6}$ & 0.109$_{6}$ & 0.939$_{3}$ \\
\bf 7 & NileTMRG & \bf 0.077$_{7}$ & 0.120$_{7}$ & 1.228$_{7}$ \\
\bf 8 & NRU-HSE & \bf 0.078$_{8}$ & 0.132$_{8}$ & 1.528$_{8}$ \\
\bf 9 & EICA & \bf 0.092$_{9}$ & 0.143$_{9}$ & 1.922$_{9}$ \\
\bf 10 & THU\_HCSI\_IDU & \bf 0.129$_{10}$ & 0.179$_{10}$ & 2.428$_{11}$ \\
\bf 11 & Amobee-C-137 & \bf 0.149$_{11}$ & 0.179$_{10}$ & 2.168$_{10}$ \\
\bf 12 & OMAM & \bf 0.164$_{12}$ & 0.204$_{12}$ & 2.790$_{12}$ \\
\bf 13 & SSK\_JNTUH & \bf 0.421$_{13}$ & 0.314$_{13}$ & 2.983$_{13}$ \\
\bf 14 & ELiRF-UPV & \bf 1.060$_{14}$ & 0.593$_{15}$ & 7.991$_{15}$ \\
\bf 15 & YNU-HPCC & \bf 1.142$_{15}$ & 0.592$_{14}$ & 7.859$_{14}$ \\
\hline
%& baseline (1 0) & 2.255 & 0.578 & 7.621\\
B1 & (0 1) & 1.518 & 0.422 & \bf 2.645\\
B2 & macro-avg on 2016 data & 0.554 & 0.423 & 6.061\\
B3 & micro-avg on 2016 data & 0.591 & 0.432 & 6.169\\
B4 & macro-avg on 2015-6 data & \bf 0.534 & \bf 0.418 & 6.000\\
B5 & micro-avg on 2015-6 data & 0.587 & 0.431 & 6.157\\
\hline
\end{tabular}
\caption{\textbf{Results for Subtask D ``Tweet quantification according to a two-point scale'', English.} The systems are ordered by their $KLD$ score (lower is better). B$x$ indicates a baseline.}
\label{table:ResultsSubtaskD:English}
\end{small}
\end{table}

\begin{table}[tbh]
\centering
\begin{small}
\renewcommand{\arraystretch}{1.0}% Tighter
\setlength{\tabcolsep}{3.3pt}
\begin{tabular}{|c|l|l|l|l|}
\hline
      \bf \# & \bf System & \multicolumn{1}{c|}{$KLD$} & \multicolumn{1}{c|}{$AE$} &\multicolumn{1}{c|}{$RAE$} \\
\hline
\bf 1 & NileTMRG & \bf 0.127$_{1}$ & 0.170$_{1}$ & 2.462$_{1}$ \\
\bf 2 & OMAM & \bf 0.202$_{2}$ & 0.238$_{2}$ & 4.835$_{2}$ \\
\bf 3 & ELiRF-UPV & \bf 1.183$_{3}$ & 0.537$_{3}$ & 11.434$_{3}$ \\
\hline
%& baseline (1 0) & 1.638 & 0.479 & 4.657\\
B1 & (0 1) & 1.518 & 0.422 & \bf 2.645\\
B2 & macro-avg on train-2017 & 0.296 & 0.322 & 6.600\\
B3 & micro-avg on train-2017 & \bf 0.295 & \bf 0.321 & 6.692\\
\hline
\end{tabular}
\caption{\textbf{Results for Subtask D ``Tweet quantification according to a two-point scale'', Arabic.} The systems are ordered by their $KLD$ score (lower is better). B$x$ indicates a baseline.}
\label{table:ResultsSubtaskD:Arabic}
\end{small}
\end{table}

\noindent \emph{funSentiment}, ranked 6th and 9th for subtasks B and C, respectively, modeled the sentiment towards the topic using the left and the right context around a topic mention in the tweet. \emph{WarwickDCS}, ranked 8th, used simple tweet-level classification, while ignoring the topic. Overall, almost all teams managed to outperform the majority class baseline for subtask B, but only two teams outperformed the \textsc{Neutral} class baseline for subtask C.

\paragraph{For Arabic} four teams participated in Subtask B and two teams in Subtask C. \emph{NileTMRG} was once again ranked first for Subtask B, with a system based on ensembles of topic-specific and topic-agnostic models.
For subtask C, \emph{OMAM} also used combinations of such models applied in succession.  
%Interestingly, they found that topic-specific models were helpful for subtask C while ignoring topics performed better for subtask B.
All teams easily outperformed the baselines for Subtask B, but only the \emph{OMAM} team managed to do so for Subtask C.

%\newpage
\begin{table}[h!]
\centering
\begin{small}
\renewcommand{\arraystretch}{1.0}% Tighter
\begin{tabular}{|c|l|l|}
\hline
      \bf \# & \bf System & \multicolumn{1}{c|}{$EMD$} \\
\hline
1 & BB\_twtr & 0.245 \\
2 & TwiSe & 0.269 \\
3 & funSentiment & 0.273 \\
4 & ELiRF-UPV & 0.306 \\
5 & NRU-HSE & 0.317 \\
6 & Amobee-C-137 & 0.345 \\
7 & OMAM & 0.350 \\
8 & Tweester & 0.365 \\
9 & THU\_HCSI\_IDU & 0.385 \\
10 & YNU-HPCC & 0.447 \\
11 & DataStories & 0.595 \\
12 & EICA & 1.461 \\
\hline
B1 & (0 0 0 1 0) & 1.123\\
B2 & macro-avg on 2016 data & 0.583\\
B3 & micro-avg on 2016 data & \bf 0.552\\
\hline
\end{tabular}
\caption{\textbf{Results for Subtask E ``Tweet quantification according to a five-point scale'', English.} The systems are ordered by their $EMD$ score (lower is better). B$x$ indicates a baseline.}
\label{table:ResultsSubtaskE:English}
\end{small}
\end{table}

\begin{table}[h!]
\centering
\begin{small}
\renewcommand{\arraystretch}{1.0}% Tighter
\begin{tabular}{|c|l|l|}
\hline
      \bf \# & \bf System & \multicolumn{1}{c|}{$EMD$} \\
\hline
1 & OMAM & 0.548 \\
2 & ELiRF-UPV & 0.564 \\
\hline
B1 & (0 0 1 0 0) & 0.458\\
B2 & macro-avg on train-2017 & \bf 0.440\\
B3 & micro-avg on train-2017 & \bf 0.440\\
\hline
\end{tabular}
\caption{\textbf{Results for Subtask E ``Tweet quantification according to a five-point scale'', Arabic.} The systems are ordered by their $EMD$ score (lower is better). B$x$ indicates a baseline.}
\label{table:ResultsSubtaskE:Arabic}
\end{small}
\end{table}

\subsection{Results for Subtasks D and E: Tweet Quantification}

Tables \ref{table:ResultsSubtaskD:English}--\ref{table:ResultsSubtaskE:Arabic} show the results for the tweet quantification subtasks.
The bottom of the tables report the result of a baseline system, B1, that assigns a prevalence of 1 to the majority class (which is the \textsc{Positive} class for subtask D, and the \textsc{WeaklyPositive}/\textsc{Neutral} class for subtask E, English/Arabic) and 0 to the other class(es). 

We further show the results for a smarter ``maximum likelihood'' baseline, which assigns to each test topic the distribution of the training tweets (the union of TRAIN, DEV, DEVTEST) across the classes. This is the ``smartest'' among the trivial policies that attempt to maximize $KLD$. 
For this baseline, for English we use for training either (\emph{i})~the 2016 data only, or (\emph{ii})~data from both 2015 and 2016; we also experiment with (\emph{i})~micro-averaging and (\emph{ii})~macro-averaging over the topics. It turns out that macro-averaging over 2015+2016 data is the strongest baseline in terms of $KLD$. For Arabic, we use the train-2017 data, and micro-averaging works better there.

\noindent There were 15 participating teams competing in Subtask D: 15 for English and 3 for Arabic (these 3 teams all participated in English). As in the other subtasks, \emph{BB\_twtr} was ranked first in English. They 
%used Deep Learning to 
achieved an improvement of .50 points absolute in KLD over the best baseline, and a .01 improvement over the next best team, \emph{DataStories}. 
%using the main scoring metric of KLD. 
For Arabic, the best team was \emph{NileTMRG}
%used Deep Learning and other classifiers 
With improvement of .17 over the best baseline and of .08 over the next best team, \emph{OMAM}. All but the last two teams in English and the last team for Arabic outperformed all baselines.

In Subtask E, there were 12 participating teams, with \emph{OMAM} and \emph{EliRF-UPV} competing for both English and Arabic. Once again, \emph{BB\_twtr} was the best for English, improving over the best baseline by .31 EMD points absolute.
Interestingly, this is the first subtask where \emph{DataStories} was not the second-ranked team.
%Instead it fell down to 11th place. 
\emph{BB\_twtr} outperformed the second-best team, \emph{TwiSe}, by .02 points. For English, all but the last two teams outperformed the baselines. However, for Arabic, none of the two participating teams could do so. 

%\subsection{Common Resources and Methods}

%In this section we describe the common trends in resources and methods across subtasks to provide a glimpse of useful tools available. Common software used includes Python (indluding sklearn and numpy), Java, Tensorflow, Weka, NLTK, Keras, Theano, and Stanford CoreNLP. The most common external datasets used were sentiment140 as a lexicon, word2vec for embeddings, and many teams gathered additional tweets using the Twitter API that were not annotated for sentiment. Deep learning including RNN, CNN, and LSTM, and word embeddings continue to be common approaches in addition to more classical supervised approaches of SVM, Logistic Regression, Random Forest, and Naive Bayes.

\subsection{User Information}

This year, we encouraged teams to explore using in their models information about the user who wrote the tweet, which can be extracted from the public user profiles of the respective Twitter users.  Participants could also try features about following relations and the structure of the social network in general, as well as could make use of other tweets by the target user when analyzing one particular tweet. Four teams tried that: \emph{SINAI}, \emph{ECNU}, \emph{TakeLab}, and \emph{OMAM}. \emph{OMAM} and \emph{TakeLab} did not find any improvements, and ultimately decided not to use any user information. \emph{ECNU} used profile information such as favorited, favorite count, retweeted, and retweet count. They ended up 15th in Subtask A.
%, which is .05 points behind the best team. 
\emph{SINAI} used the last 200 tweets from the person's timeline. They ranked 12th in Subtask B.
%, at .06 points behind the best team. 
They generated a user model from the timeline of a given target user. They built a general SVM model on word2vec embeddings. Then, for each user in the test set, they downloaded the last 200 tweets published by the user and classified their sentiment using that SVM classifier. If the classified user tweets achieved an accuracy above a threshold (0.7), the user model was applied on the authored tweets from the test set. If not, the general SVM model was used. 

\noindent It is difficult to judge whether and by how much user information could help the best approaches as they did not try to use such information.
However, we believe that building and using a Twitter user profile is a promising research direction, and that participants should learn how to make this work in the future. Thus, we would like to encourage more teams to try to explore using this information. 
We would also like to provide more user information such as age and gender, which we can predict automatically~\cite{rosenthal2016social}, when it is not directly available from the user profile.  
Another promising direction is to make use of ``conversations'' in Twitter, i.e., take into account the replies to tweets in Twitter.
For example, previous work \cite{vanzo-croce-basili:2014:Coling} has shown that it is beneficial to model the polarity detection problem as a sequential classification task over streams of tweets, where the stream is a ``conversation'' on Twitter containing tweets, replies to these tweets, replies to these replies, etc.

\section{Conclusion and Future Work}
\label{sec:conclusion}

Sentiment Analysis in Twitter continues to be a very popular task, attracting 48 teams this year. The task provides immense value to the sentiment community by providing a large accessible benchmark dataset containing over 70,000 tweets across two languages for researchers to evaluate and compare their method to the state of the art. This year, we introduced a new language for the first time and also encouraged the use of user information. These additions drew new participants and ideas to the task. The Arabic tasks drew nine participants and four teams took advantage of user information. Although a respectable amount of participants for its inaugural year, further exploration into both of these areas would be useful in the future, such as collecting more training data for Arabic and encouraging the use of cross-lingual training data. In the future, we would like to include exploring additional languages, providing further user information, and other related tasks such as irony and emotion detection. Finally, deep learning continues to be popular and employed by the state of the art approaches. We expect this trend to continue in sentiment analysis research, but also look forward to new innovative ideas that are discovered. 

%\newpage
\begin{table*}[h!]
\centering
\begin{scriptsize}
\setlength{\tabcolsep}{3pt}
\begin{tabular}{|l|p{4.5cm}|p{1.8cm} | lllll | lllll |p{3.8cm}|}
\hline
\multirow{2}{*}{\bf Team ID}	&	\multirow{2}{*}{\bf Affiliation}	&	\multirow{2}{*}{\bf Country}	&	\multicolumn{10}{c|}{\bf Subtasks}		&	\multirow{2}{*}{\bf Paper} \\
	&		&		&	\multicolumn{5}{c|}{English} &	\multicolumn{5}{c|}{Arabic} & \\ \hline
    \hline
Adullam	&	Korea University	&	South Korea	&	A	&		&		&		&		&		&		&		&		&		&	\cite{SemEval:2017:task4:adullam}	\\
\hline
Amobee C-137	&	Amobee	&	USA	&	A	&	B	&	C	&	D	&	E	&		&		&		&		&		&	\cite{SemEval:2017:task4:Amobee}\\
\hline
ASA	&	Al-Imam Muhammad Ibn Saud Islamic University.	&	Saudi Arabia	&		&		&		&		&		&		&	B	&		&		&		&	N/A	\\
\hline
Avid	&	N/A	&	N/A	&	A	&		&		&		&		& 		&		&		&		&		&	N/A	\\
\hline
BB\_twtr	&	Bloomberg	&	USA	&	A	&	B	&	C	&	D	&	E	&		&		&		&		&		&	\cite{SemEval:2017:task4:BB_twtr}\\
\hline
BUSEM 	&	Bogazici University	&	Turkey	&	A	&		&		&		&		&		&		&		&		&		&	\cite{SemEval:2017:task4:busem}\\
\hline
CrystalNest	&	Institute of High Performance Computing (IHPC) 	&	Singapore	&	A	&	B	&	C	&	D	&		&		&		&		&		&		&	\cite{SemEval:2017:task4:CrystalNest}\\
\hline
DataStories	&	Data Science Lab at University of Piraeus	&	Greece	&	A	&	B	&	C	&	D	&	E	&		&		&		&		&		&		\cite{SemEval:2017:task4:datastories} \\
\hline
deepSA	&	National Sun Yat-sen University	&	Taiwan	&	A	&		&		&		&		&		&		&		&		&		&	\cite{SemEval:2017:task4:deepSA}\\
\hline
diegoref	&	N/A	&	N/A	&	A	&		&		&		&		& 		&		&		&		&		&	N/A	\\
\hline
DUTH	&	Democritus University of Thrace	&	Greece	&	A	&	B	&	C	&		&		&		&		&		&		&		&	\cite{SemEval:2017:task4:duth}\\
\hline
ECNU	&	East China Normal University	&	China	&	A	&		&		&		&		&		&		&		&		&		&	\cite{SemEval:2017:task4:ecnu} \\
\hline
EICA	&	East China Normal University	&	China	&	A	&	B	&	C	&	D	&	E	&		&		&		&		&		&	\cite{SemEval:2017:task4:eica} \\
\hline
ej-sa-2017	&	University of Evora	&	Portugal	&	A	&	B	&		&		&		&		&		&		&		&		&	\cite{SemEval:2017:task4:ej-sa-2017} \\
\hline
ELiRF-UPV	&	Universitat Polit\'{e}cnica de Val\'{e}ncia	&	Spain	&	A	&	B	&	C	&	D	&	E	&	A	&	B	&	C	&	D	&	E	&	\cite{SemEval:2017:task4:elirf-upv} \\
\hline
funSentiment	&	Thomson Reuters	&	USA	&			&	B	&	C	&	D	&	E	&	& 		&		&		&		&	\cite{SemEval:2017:task4:funsentiment}	\\
\hline
HLP@UPENN	&	University of Pennsylvania	&	USA	&	A	&		&		&		&		&	A	&		&		&		&		&	\cite{SemEval:2017:task4:hlp@upenn} \\
\hline
INGEOTEC	&	CONACYT-INFOTEC/CENTROGEO	&	Mexico	&	A	&		&		&		&		&	A	&		&		&		&		&	\cite{SemEval:2017:task4:ingeotec} \\
\hline
LIA	&	LIA	&	France	&	A	&		&		&		&		&		&		&		&		&		&	\cite{SemEval:2017:task4:lia} \\
\hline
LSIS	&	Aix-Marseille University	&	France	&	A	&		&		&		&		&	A	&		&		&		&		&	\cite{SemEval:2017:task4:lsis}	\\
\hline
MI\&T Lab	&	Harbin Institute of Technology	&	China	&	A	&		&		&		&		&		&		&		&		&		&	\cite{SemEval:2017:task4:mitlab} \\
\hline
Neverland-THU	&	N/A	&	N/A	&	A	&		&		&		&		& 		&		&		&		&		&	N/A	\\
\hline
NILC-USP	&	Institute of Mathematics and Computer Science, University of São Paulo	&	Brazil	&	A	&		&		&		&		&		&		&		&		&		&	\cite{SemEval:2017:task4:nilc-usp}\\
\hline
NileTMRG	&	Nile University	&	Egypt	&	A	&	B	&		&	D	&		&	A	&	B	&		&	D	&		&	\cite{SemEval:2017:task4:niletmrg} \\
\hline
NNEMBs	&	Peking University	&	China	&	A	&		&		&		&		&		&		&		&		&		&	\cite{SemEval:2017:task4:nnembs} \\
\hline
NRU-HSE	&	National Research University Higher School of Economics	&	Russia	&		&	B	&	C	&	D	&	E	&		&		&		&		&		&	\cite{SemEval:2017:task4:nru-hse} \\
\hline
OMAM	&	American University of Beirut, Universiti Teknologi Malaysia, Cairo University, New York University Abu Dhabi, Qatar University	&	Egypt, Lebanon, Malaysia, Qatar, United Arab Emirates	&	A	&	B	&	C	&	D	&	E	&	A	&	B	&	C	&	D	&	E	&	\cite{SemEval:2017:task4:omam,SemEval:2017:task4:omam2} \\
\hline
QUB	&	Queen's University Belfast	&	Ireland	&	A	&		&		&		&		&		&		&		&		&		&		\\
\hline
senti17	&	Lip6, UPMC	&	France	&	A	&		&		&		&		&		&		&		&		&		&	\cite{SemEval:2017:task4:Senti17}\\
\hline
SentiME++	&	EURECOM	&	France	&	A	&		&		&		&		&		&		&		&		&		&	\cite{SemEval:2017:task4:SentiME++} \\
\hline
SINAI	&	Universidad de Ja\'{e}n	&	Spain	&		&	B  &		&		&		&		&		&		&		&		&	\cite{SemEval:2017:task4:sinai} \\
\hline
SiTAKA	&	iTAKA, Universitat Rovira i Virgili; Hodeidah University	&	Spain, Yemen	&	A	&		&		&		&		&	A	&		&		&		&		&	\cite{SemEval:2017:task4:SiTAKA} \\
\hline
SSK\_JNTUH	&	J.N.T.U.H College of Engg Jagtial and BVRIT Hyderabad College of Engineering for Women	&	India	&		&	B	&		&	D	&		&		&		&		&		&		&	N/A	\\
\hline
SSN\_MLRG1	&	Department of CSE, SSN College of Engineering	&	India	&	A	&	B	&	C	&		&		&		&		&		&		&		&	\cite{SemEval:2017:task4:ssn_mlrg1} \\
\hline
TakeLab	&	TakeLab, University of Zagreb	&	Croatia	&	A	&	B	&		&	D	&		&		&		&		&		&		&	\cite{SemEval:2017:task4:takelab} \\
\hline
THU\_HCSI\_IDU	&	Human Computer Speech Interaction Research Group, Tsinghua University	&	China	&		&		&		&	D	&	E	&		&		&		&		&		&		\\
\hline
Ti-Senti	&	N/A	&	N/A	&	A	&	B	&		&		&		&		&		&		&		&		&	N/A \\
\hline
TM-Gist	&	N/A	&		&		&	B	&		&		&		&		&		&		&		&		&	N/A	\\
\hline
TopicThunder	&	N/A	&	N/A	&		&	B	&		&		&		&		&		&		&		&		&	N/A \\
\hline
TSA-INF	&	Infosys Limited	&	India	&	A	&		&		&		&		&		&		&		&		&		&	\cite{SemEval:2017:task4:tsa-inf}	\\
\hline
Tw-StAR	&	Selcuk University, Université Libre de Bruxelles (ULB)	&	Belgium, Turkey	&		&		&		&		&		&	A	&		&		&		&		&	\cite{SemEval:2017:task4:1w-StAr} \\
\hline
Tweester	&	National Technical University of Athens, University of Athens, ``Athena'' Research and Innovation Center, Signal Analysis and Interpretation Laboratory (SAIL), USC	&	Greece, USA	&	A	&	B	&	C	&	D	&	E	&		&		&		&		&		&	\cite{SemEval:2017:task4:Tweester} \\
\hline
TwiSe	&	University of Grenoble-Alps	&	France	&		&		&	C	&		&	E	&		&		&		&		&		&	\cite{SemEval:2017:task4:twise} \\
\hline
UCSC-NLP	&	Catholic University of the Most Holy Conception	&	Chile	&	A	&		&		&		&		&		&		&		&		&		&	\cite{SemEval:2017:task4:ucsc-nlp}\\
\hline
WarwickDCS	&	Department of Computer Science, University of Warwick	&	UK	&	A	&	B	&		&		&		&		&		&		&		&		&	N/A	\\
\hline
XJSA	&	Xi'an JiaoTong University 	&	China	&	A	&		&		&		&		&		&		&		&		&		&	\cite{SemEval:2017:task4:xjsa} \\
\hline
YNU-HPCC	&	Yunnan University	&	China	&	A	&	B	&	C	&	D	&	E	&		&		&		&		&		&	\cite{SemEval:2017:task4:ynu-hpcc} \\
\hline
YNUDLG	&	Yunnan University	&	China	&	A	&	B	&	C	&		&		&		&		&		&		&		&	\cite{SemEval:2017:task4:ynu-1510} \\
\hline
\multicolumn{3}{l}{\bf TOTAL}				&	\bf 38	&	\bf 23	&	\bf 15	&	\bf 15	&	\multicolumn{1}{c}{\bf 12}	&	\bf 8	&	\bf 4	&	\bf 2	&	\bf 3	&	\multicolumn{1}{c}{\bf 2}	&	\multicolumn{1}{c}{ } \\
\end{tabular}
\caption{Alphabetical list of the participating teams, their affiliation, country, the subtasks they participated in, and the system description paper that they contributed to SemEval-2017. Teams whose \emph{Affiliation} column is typeset on more than one row include researchers from different institutions, which have collaborated to build a joint system submission.  An \emph{N/A} entry for the \emph{Paper} column indicates that the team did not contribute a system description paper. Finally, the last row gives statistics about the total number of system submissions for each subtask.} 
\label{table:Affiliations}
\end{scriptsize}
\end{table*}
\newpage

% increase data for Arabic
% encourage cross-lingual analysis
% introduce a new language (perhaps Spanish)

% We expect the quest for more interesting formulations of the general sentiment analysis task to continue. We see SemEval as the engine of this innovation, as it not only does head-to-head comparisons, but also creates databases and tools that enable follow-up research for many years afterwards.

\bibliography{bib,participants}
\bibliographystyle{acl_natbib}

%\appendix

%\section{Supplemental Material}
%\label{sec:supplemental}

%\section{Multiple Appendices}

\end{document}